\documentclass[10pt,twocolumn,letterpaper]{article}

\usepackage[pagenumbers]{cvpr} 

\usepackage{graphicx}
\usepackage{amsmath}
\usepackage{amssymb}
\usepackage{booktabs}

\usepackage[pagebackref,breaklinks,colorlinks]{hyperref}

\usepackage[capitalize]{cleveref}
\crefname{section}{Sec.}{Secs.}
\Crefname{section}{Section}{Sections}
\Crefname{table}{Table}{Tables}
\crefname{table}{Tab.}{Tabs.}

\def\cvprPaperID{9938} 
\def\confName{CVPR}
\def\confYear{2022}

\begin{document}

\title{Semantic Segmentation In-the-Wild Without Seeing Any Segmentation Examples}

\author{Nir Zabari \quad Yedid Hoshen\\
School of Computer Science and Engineering \\
The Hebrew University of Jerusalem, Israel \\
}

\maketitle
\begin{abstract}
Semantic segmentation is a key computer vision task that has been actively researched for decades. In recent years, supervised methods have reached unprecedented accuracy, however they require many pixel-level annotations for every new class category which is very time-consuming and expensive. Additionally, the ability of current semantic segmentation networks to handle a large number of categories is limited. That means that images containing rare class categories are unlikely to be well segmented by current methods. 
In this paper we propose a novel approach for creating semantic segmentation masks for every object, without the need for training segmentation networks or seeing any segmentation masks. Our method takes as input the image-level labels of the class categories present in the image; they can be obtained automatically or manually. We utilize a vision-language embedding model (specifically CLIP) to create a rough segmentation map for each class, using model interpretability methods. We refine the maps using a test-time augmentation technique. The output of this stage provides pixel-level pseudo-labels, instead of the manual pixel-level labels required by supervised methods. Given the pseudo-labels, we utilize single-image segmentation techniques to obtain high-quality output segmentation masks. Our method is shown quantitatively and qualitatively to outperform methods that use a similar amount of supervision. Our results are particularly remarkable for images containing rare categories.

\end{abstract}
\section{Introduction}
\label{sec:intro}
%




\begin{figure*}[t]
\begin{center}
\includegraphics[width=\linewidth]{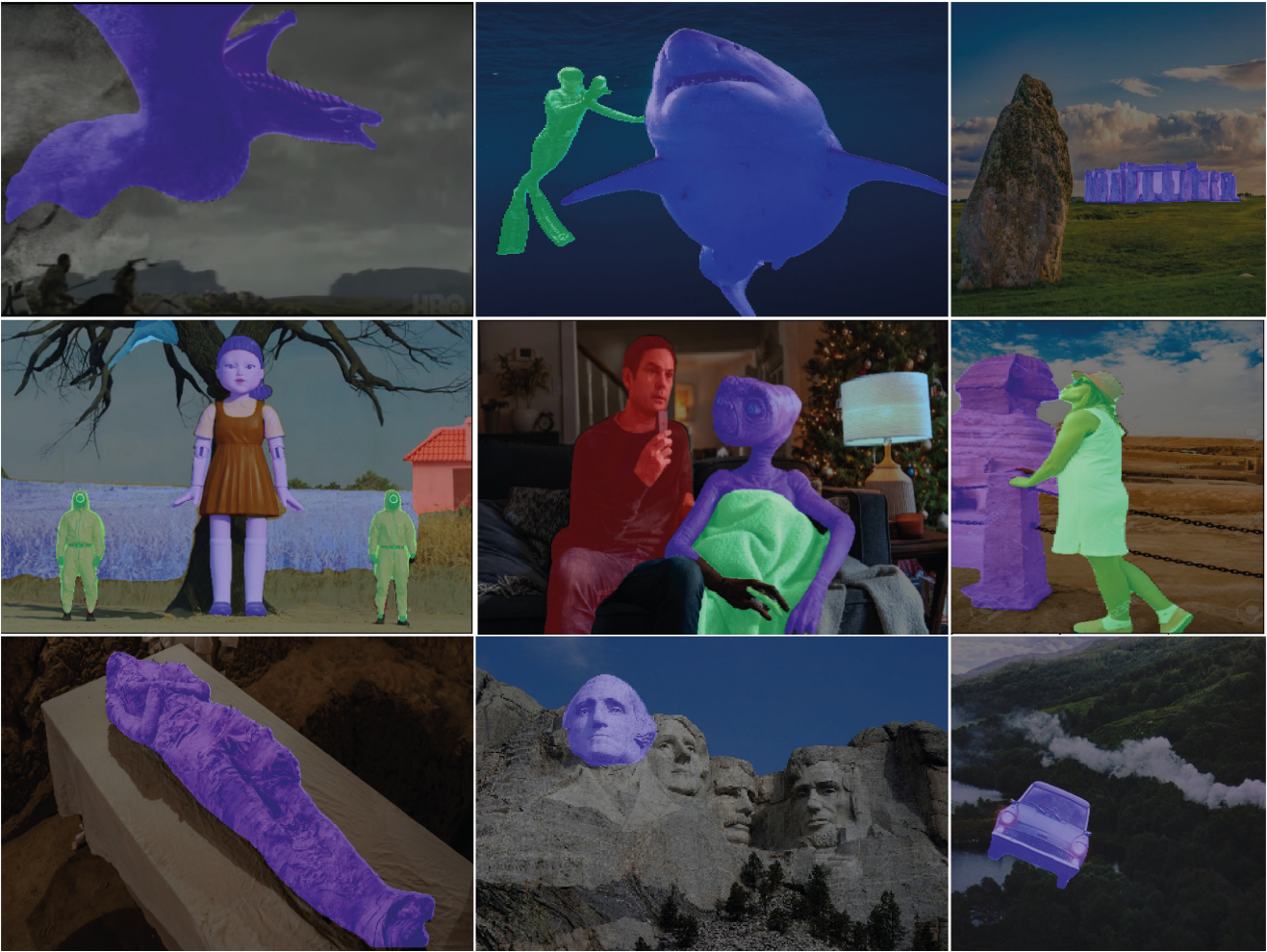}
\end{center}
\caption{
Results of our method on real-world images from rare class categories. As inputs, our algorithm receives an image along with text-prompts describing the classes that we want to segment. A language-vision model is distilled by generating a relevance map in relation to each prompt category. Further refinement is performed through test time augmentations. Next, the relevance maps are fed into a single image segmentation algorithm, which transforms the relevance maps into a high-quality segmentation.  
}
\label{fig:cool_results}
\end{figure*}
The task of semantic segmentation, which involves assigning a class category to each pixel of an image, has been evolved recently by deep neural networks trained on large-scale annotated datasets \cite{SEG_BASE_Chen2018DeepLabSI, SEG_BASE_Huang2019CCNetCA, SEG_BASE_Wang2021DeepHR}. Although this progress is very impressive, such systems are mainly useful for common semantic categories where large annotated datasets are available. However, in many real world scenarios, segmentation is required for rare class categories for which training data are unavailable. Annotation of new data for every novel category is expensive, time-consuming and impractical in most cases. Moreover, current supervised methods do not scale well to work on imbalanced datasets of large vocabularies, where common categories have many examples, while the rare categories follow a long-tailed distribution \cite{LVIS_Gupta2019LVISAD, LVIS_Wang2020TheDI}.

When a few annotated examples of a rare class are available, different paradigms have been proposed, including Few-Shot Semantic Segmentation (FSSS) and Weakly Supervised Semantic Segmentation (WSSS). The two settings differ in the expected supervision. FSSS requires a few pixel-annotated images containing the rare category, while WSSS requires very coarse supervision e.g. image level labels, bounding boxes, or scribbles but on many images. These methods cannot be applied in real-world settings where such supervision is unavailable. In order to operate in settings with no labels, Zero Shot Semantic Segmentation (ZSSS) methods were proposed, but current methods are not yet accurate enough for many use cases when applied to rare classes. 

We propose a novel method for segmenting objects that does not require new data annotation, is fast (10 second to two minutes) and achieves high accuracy results. Our method utilizes CLIP\cite{CLIPRadford2021LearningTV}, a recently developed vision-language model, which embeds visual and linguistic concepts in a shared space through large-scale contrastive learning on web data. Our method uses the documented zero-shot abilities of CLIP. Given a CLIP model and text prompts describing the semantic classes present in the image, our method first creates a set of per-category relevance map of the image. The mapping between the text prompts to per-category relevance maps is performed by utilizing a recently developed transformer interpretation method \cite{Chefer_2021_CVPR}. Since the relevance maps generated by explanation methods are noisy and inaccurate, we use custom test-time augmentations to refine the relevance maps. We then combine the individual per-category maps into a multi-class relevance map of the image. The relevance maps, which was obtained in a zero-shot manner, is already quite accurate and is used as the input to single-image segmentation methods. We present results on an unsupervised single image clustering technique that we augment with our refined relevance maps as weak pseudo supervision. We also use our method as the input to a supervised single-image supervision method, where we replace the expected supervision by our hard segmentation map. 

Our method is extensively evaluated and ablated. We quantitatively evaluate our method on standard segmentation benchmarks and demonstrate that it performs better than current methods that use a similar amount of supervision. The ability of our method to operate on very rare and unique images is clearly demonstrated (see \cref{fig:cool_results} and \cref{fig:landmark_seg}). Finally, we provide an ablation study demonstrating the importance of the different components of the method. Our main contributions are:
\begin{enumerate}
    \item Proposing  a new method for segmenting any kind of object without requiring pixel-level annotations or multiple sample images.
    \item Proposing the idea of using pre-trained language-vision networks for semantic segmentation of any object category.
    \item An extensive evaluation showing better performance than other methods using the same level of supervision, with impressive results on rare objects.
\end{enumerate}

\section{Related works}
\label{sec:related}
Most current semantic segmentation models \cite{SEG_BASE_Chen2018DeepLabSI, SEG_BASE_Huang2019CCNetCA, SEG_BASE_Wang2021DeepHR, SEG_CLOSE_Yuan2020ObjectContextualRF, SEG_CLOSE_Zhang2021DCNASDC} classify pixels into a fixed set of closed categories for which ample supervision is provided. weakly-supervised semantic segmentation, few-shot semantic segmentation, and zero-shot semantic segmentation extend these methods by generating segmentation masks for semantic classes with simpler-to-obtain annotations, a limited number of annotated images or no supervision, in order to reduce the requirement for pixel annotations. These methods have evolved rapidly in recent years.

\textbf{Weakly Supervised Semantic-Segmentation (WSSS).} These approaches use weak forms of supervision such as image labels \cite{WSSS_LABELS_Pathak2015ConstrainedCN, WSSS_LABELS_Hong2017WeaklySS, WSSS_LABELS_Araslanov2020SingleStageSS}, bounding boxes \cite{WSSS_BOX_Dai2015BoxSupEB, WSSS_BOX_Kervadec2020BoundingBF}, and scribbles \cite{WSSS_SCRIBBLE_Lin2016ScribbleSupSC, WSSS_SCRIBBLE_Zhang2020WeaklySupervisedSO}. Image-level labels are probably the most popular form of weak supervision, due to their simplicity and the possibility of obtaining them from public datasets or web data. 
A typical WSSS pipeline begins with generating a pseudo mask, followed by training a new semantic segmentation network \cite{WSSS_PIPE_Lee2021RailroadIN}. Interpretability techniques such as CAM \cite{CAM_Selvaraju2019GradCAMVE} are often used to infer incomplete pixel-level annotations automatically. These masks are often inaccurate as interpretability methods usually generate pseudo masks that highlight only the most discriminative parts (e.g. highlight the railroad when classifying a train). They therefore cannot generate a complete semantic map, but only be used as a seed for segmentation. There have been several proposed methods for improving the segmentation masks generated from CAM, such as PSA \cite{WSSS_PSA_Ahn_2018_CVPR} and IRN  \cite{WSSS_IRN_Ahn_2019_CVPR} which use boundary information by calculating affinities between pixels. Other methods \cite{WSSS_SEED_Kolesnikov2016SeedEA, WSSS_SEED_Huang2018WeaklySupervisedSS} start with seeds and build up to the object boundary iteratively. While our method also uses image-level annotations, it differs from the above techniques as no training images are required.

\textbf{Few-Shot Semantic-Segmentation (FSSS).}
Following the success of few-shot and zero-shot in classification \cite{CLIPRadford2021LearningTV,ZSClassification_Narayan2020LatentEF} and object detection \cite{ZSOD_Bansal2018ZeroShotOD, ZSOD_Rahman2018ZeroShotOD, ZSOD_Rahman2020ImprovedVA}, few-shot and zero-shot semantic segmentation methods emerged. FSSS typically require a small number of segmented training images, usually fewer than ten. The annotated examples serve as the support set, from which later segmentation masks are generated for the input image, called a query set. \cite{FSSS_Zhang2020SGOneSG} uses masked average pooling to extract foreground and background information. \cite{FSSS_Dong2018FewShotSS} leverages metric-learning for few-shot segmentation. \cite{FSSS_Rakelly2018ConditionalNF} employed an encoder-decoder architecture with auxiliary conditional encoder branches that concatenated features from both query and support images to feed the decoder.

\textbf{Zero-Shot Semantic-Segmentation (ZSSS).} While FSSS methods used zero-shot learning to recognition in the image-level, these methods use zero-shot learning at the level of individual pixels. These approaches learn at training time to classify pixels into a closed set of class categories, and at test-time apply these models on never-seen class categories. \cite{ZSSS_nips_bucher2019zero} used a state of the art segmentation network (DeepLab), and propagate information of unseen classes to pixel embeddings using Word2Vec, together with self-training. \cite{ZSSS_CAP_Tian2020Cap2SegIS} leveraged image captions in order to segment unknown classes. \cite{ZSSS_StructureLi2020ConsistentSR}  learns a generator to produce visual features from semantic word embeddings, similar to \cite{ZSSS_nips_bucher2019zero}, but it alternated between generating “good features”, while maintaining the structural-relations between  categories in the text latent space, as before. 

\textbf{Shared Language-Vision Latent space. } Recently there have been multiple works which have proposed learning shared text and image embeddings in the same latent space
\cite{VL_Desai2021VirTexLV,VL_Li2020UnicoderVLAU,VL_Sariyildiz2020LearningVR,VL_Tan2019LXMERTLC, CLIPRadford2021LearningTV}. 
CLIP (Contrastive Language Image Pre-Training) uses 400M image-caption pairs to assess the similarity between a text and an image, as a pre-training task. A variety of computer vision tasks have been demonstrated to benefit from CLIP model embeddings, which demonstrated robustness and generalization to a wide range of visual concepts. Some notable applications which  draw on the CLIP's shared embedding space are \cite{CLIP_APP_StyleClip_Patashnik2021StyleCLIPTM} used the latent space of StyleGan to  generate images based on a text description. \cite{CLIP_APP_domain_adaptation_Gal2021StyleGANNADACD} adapts an image generator to other domains with zero examples. Using CLIP's latent space, \cite{CLIP_app_captioning_Galatolo2021GeneratingIF} generated an image from its caption, and vice versa. \cite{CLIP_APP_vid_retrievalLuo2021CLIP4ClipAE}  used CLIP for video retrieval.









\section{Method}

\begin{figure}[t]
\begin{center}
\includegraphics[width=.99\linewidth]{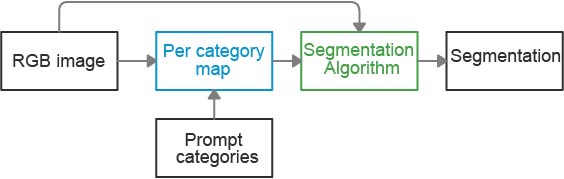}
\end{center}
\caption{
\textbf{Segmentation Pipeline} - A high-level description of our method. We receive an RGB image to process and user-suggested categories for segmentation. This assumes the prompt categories actually appear in the image. By using an interpretability method, we produce a relevance map with respect to each prompt category. Afterwards, we employ a segmentation algorithm (e.g. clustering or interactive segmentation algorithms) that is able to segment the image and leverage the pseudo labels induced by the relevance maps. Our final output is a segmentation map.
}
\label{fig:pipeline_abstract}
\end{figure}
\begin{figure}[t]
\begin{center}
\includegraphics[width=.99\linewidth]{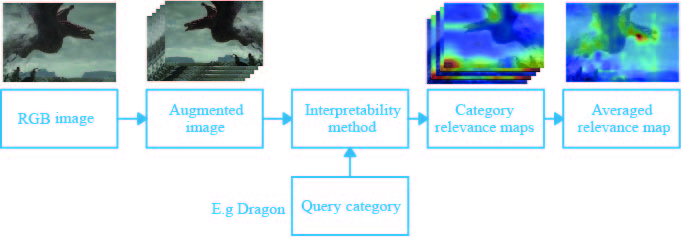}
\end{center}
\caption{
\textbf{Relevance Maps Generation} - A sketch of the relevance map generation module, which can generate a relevance map for possibly any type of object. We utilize an interpretability method, a language-vision pre-trained model using TTA techniques, to obtain a refined relevance map.
}
\label{fig:heatmap_creation}
\end{figure}

The input to our method is an RGB image $I$ of shape  $(h,w)$ and a list of $K$ categories, each given by a text prompt $(T_1,T_2,...,T_K)$. The output of our method is a segmentation mask $S(I) := S \in \{0,..,K\}^{(h,w)}$. A pixel is denoted $p=(x,y)$. Background pixels $p$ are denoted by $S_p=0$, other categories $i$ are denoted by $S_p=i$.
Our method consists of two stages as shown in \cref{fig:pipeline_abstract}. In the first stage (see \cref{fig:heatmap_creation}), we compute a relevance map for each prompt category $\mathcal{C}$, where a larger value for a pixel indicates a higher likelihood of belonging to category  $\mathcal{C}$. A relevance map is created for each category given its text prompt, by utilizing a pre-trained vision-language model (here we use CLIP). The map is created by finding the image regions that explain the decision made by CLIP. To increase the map accuracy to a high-enough accuracy so they can serve as a source of synthetic supervision, we employ test-time-augmentation (TTA) techniques to obtain the final relevance map.\
In the second stage, we convert the image from the averaged relevance map image $SS\in [0,1]^k$, where $SS_p(\mathcal{C})$ for $1\leq\mathcal{C}\leq k$ is indication of the pixel $p$ to be of category $\mathcal{C}$, to the final segmentation image $S \in \{0,..,k\}^{(h,w)}$. The transition from the relevance maps to the segmentation image can be achieved in multiple ways: weakly-supervised clustering by scribble, interactive segmentation techniques, active contours models, etc. The synthetic supervision is generated from the map using stochastic pixel sampling, and is then fed to downstream methods. We experimented with weakly supervised clustering and interactive segmentation models.  

\subsection{Relevance Map Generation}
\textbf{Relevance Map Mining.} Our observation is that we can leverage large pre-trained language-vision models to segment any object.  These models were previously trained on very large datasets of image-caption pairs and have demonstrated great performance for zero-shot classification. We suggest that interpretability techniques can be used to infer dense class labels for rare class categories. For each prompt category, we created a relevance map using an interpretability technique. We denote this by a function $M$, that takes as input an image $I$ and prompt $T_{\mathcal{C}}$ and returns a relevance map $M(I, T_{\mathcal{C}})$. Test-time augmentation techniques are used to enhance the maps since they suffer from noise, tend to extend beyond object boundaries, and can be incomplete or imprecise. Hence, for each prompt category $\mathcal{C}$ and an input image $I$, a set of random image augmentation functions $f_1,f_2..f_V$, $V$ views are generated: 
\begin{equation}
 \textit{views} = \{f_v(I)\}_{v=1}^{V}
\end{equation}

We then average our maps to reduce noise and obtain the averaged relevance maps for category $\mathcal{C}$:
\begin{equation}
    RM(\mathcal{C}) = \mathbb{E} [ M(f_v(I), T_{\mathcal{C}}) ] 
\end{equation}

\begin{figure}[t]
\begin{center}
\includegraphics[width=.99\linewidth]{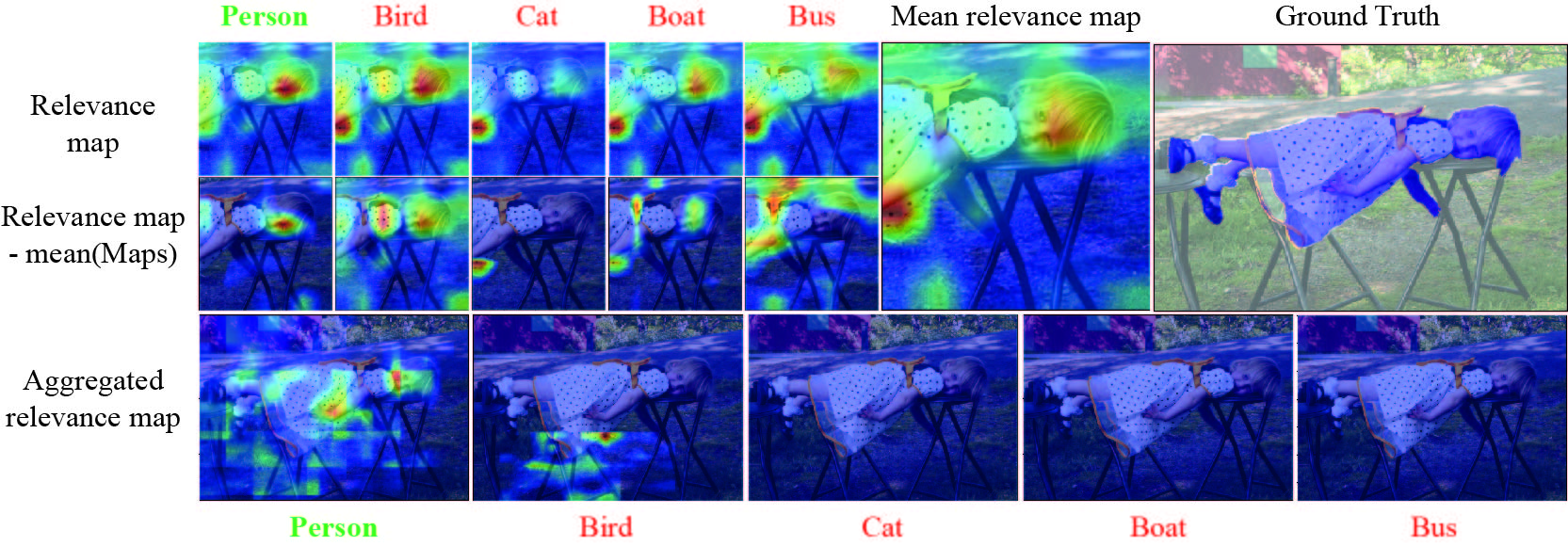}
\end{center}
\caption{
\textbf{Crop Augmentation} - Crops are selected from a grid overlaid on the image and relevance maps are calculated for each crop in relation to the prompt category. As crops overlap, each pixel's relevance value is averaged across different crops. In order to remove the noise and artifacts produced by the interpretability method, we subtracted the mean of the relevance maps of both the query category (e.g. Person, in green) and the distractor categories (Bird, Cat, Boat, and Bus, in red). 
Then, the probability of each class category is obtained using CLIP, which allows us to analyze only relevant labels, even if we do not know what the labels are. When our prompt category class probability is greater than a specified threshold (we used $\mathbb{P}(class) > 0.3$), we add the crop's relevance maps to an aggregated relevance map of the entire image.  Finally, the aggregated relevance map is normalized to the range $[0,1]$, resulting in the final crop view. }
\label{fig:reduce_noise_crop}
\end{figure}

\textbf{Image views.} We denote different image transformations as views. The 4 views that we use are: the original image, horizontal flipping, randomly changing the image contrast and a random crop of the image. Due to the effectiveness of the cropping operation, we compute results for multiple image crops and then aggregate them back into the complete relevance map (see \cref{fig:reduce_noise_crop}). \\

\textbf{Relevance Map Refinement.} Despite averaging over multiple augmentations, the resulting maps are often not specific enough. Consider an image with a person and a doll. A query for the label "person" will result in high values in the generated maps for both the person and doll pixels and vice versa. To reduce this inaccuracy, we calibrate the maps by computing relevance maps for several distractor prompts. We compute the average relevance map for the distractor classes and remove the mean of the distractor maps from each of our per-category maps. This significantly improves the specificity of the maps, as demonstrated in \cref{fig:reduce_noise_crop} .

 \label{relevanece_maps_to_segmentation}
\subsection{From View-Averaged Relevance Maps to Segmentation}
The obtained view-averaged per-category relevance maps have a much higher quality than the naive outputs of the interpretability technique. However, they are still not precise enough, e.g. segmentations typically spill over beyond the object boundaries. We would like to transform the multiple maps into a single segmentation, where each pixel is classified to a semantic label, i.e an integer number $ p \in \{0,..,k\}$.

We propose to combine our maps with existing single-image segmentation methods. We investigate both the combination with unsupervised methods (therefore increasing their performance) or adaptation of supervised methods by replacing their pixel-supervision by our maps as "pseudo-supervision":

\textbf{Unsupervised Clustering \cite{Kim2020UnsupervisedLO}.} As a baseline, we used the method proposed by \cite{Kim2020UnsupervisedLO}, which reframes image segmentation as a pixel clustering problem. An image is optimized with a small fully convolutional network to optimize both the feature representation and pixel labeling together. Using this technique, pixels of similar features will be assigned the same label, while still being being segmented continuously due to a continuity prior. We inject supervision to this technique, by using \textit{Stochastic Pixel Sampling} (see below) and adding these "pseudo labels" as an additional classification loss. The network operates directly on deep features of a pretrained network. We stop the clustering process after a given number of iterations, or when we the minimum number of classes is reached (usually the number of prompt categories + 1). More details can be found in the SM.

\textbf{Interactive Segmentation.} This segmentation paradigm guides the results by interaction, where users provide boundary seeds, regions of interest (ROI), bounding boxes, regions-seeds i.e foreground points and background points. We employed  interactive segmentation \cite{interactive_seg_Sofiiuk2021RevivingIT}, that takes as an input the region-seeds. The foreground and background points were generated on the fly using our \textit{Stochastic Pixel Sampling}.

\textbf{Stochastic Pixel Sampling.} The simplest option of transforming mean per-category relevance maps into coarse segmentation is by taking the maximum class of every pixel, thresholded with the background probability. We found that better downstream segmentation can be obtained by stochastic pixel sampling.  We propose an auxiliary operation to stochastically sample pixels of the prompt categories or of the background. We sample pixels using the following probability (where $\tau$ is the temperature):
\begin{equation}
    \mathbb{P}(p) \sim Softmax\left(\frac{RM(\mathcal{C})}{\tau} \right)(p)
\end{equation}

\begin{figure}[t]
\begin{center}
\includegraphics[width=.99\linewidth]{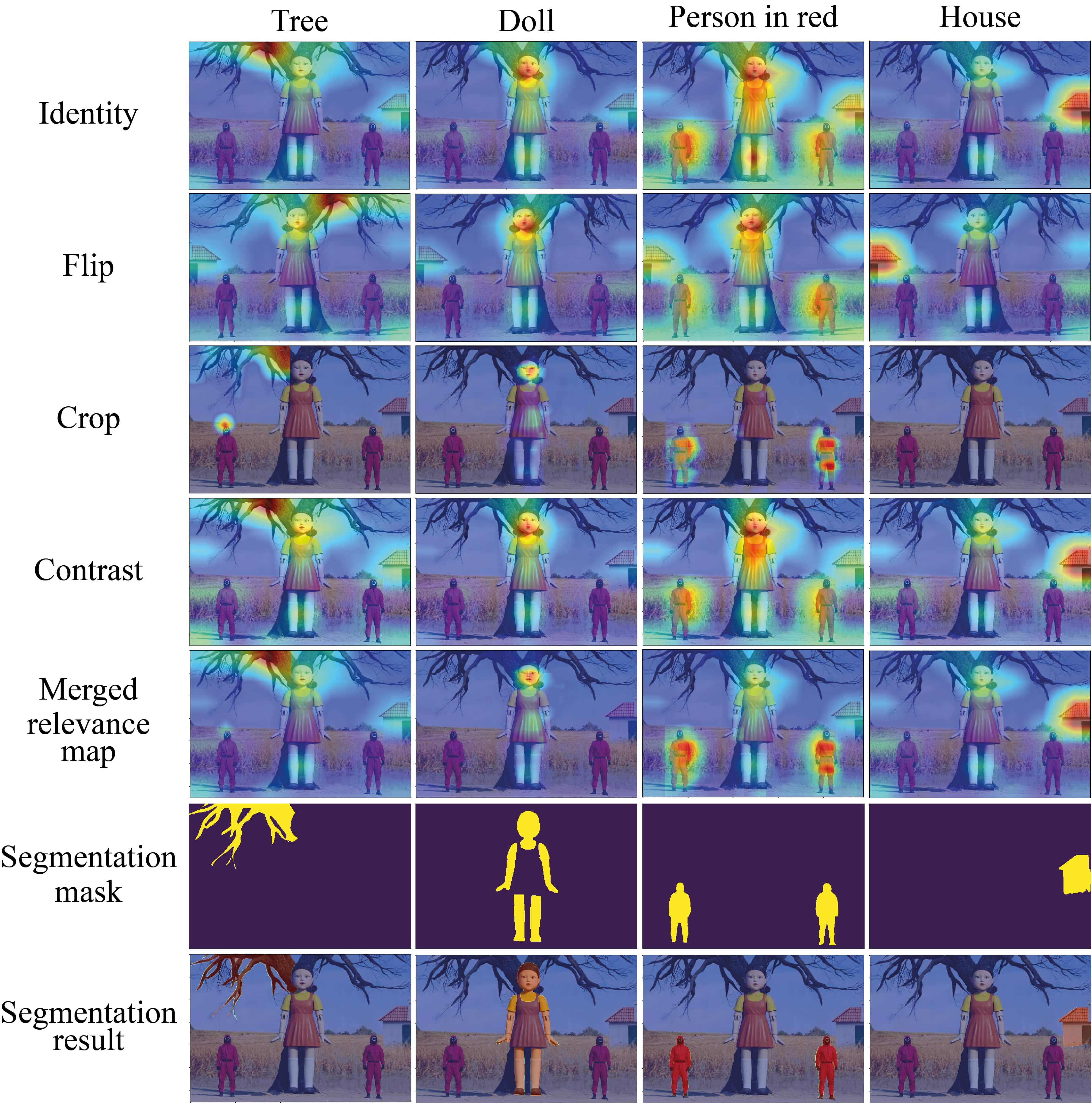}
\end{center}
\caption{
\textbf{Obtaining clean relevance maps via Test Time Augmentation} - Over a representative set of augmentations, a relevance map is created for each prompt category. By averaging all the  relevance maps for each augmentation view, a  more subtle map can be obtained for each category. While crop augmentation generally provides a finer map, other augmentations that work over the whole image usually produce a coarser map. Segmentation depends heavily on the map quality, which tends to be noisy.
}
\label{fig:augmentation_reduction}
\end{figure}

\section{Experiments}

\begin{figure*}[t] 

\begin{center}
\includegraphics[width=\linewidth]{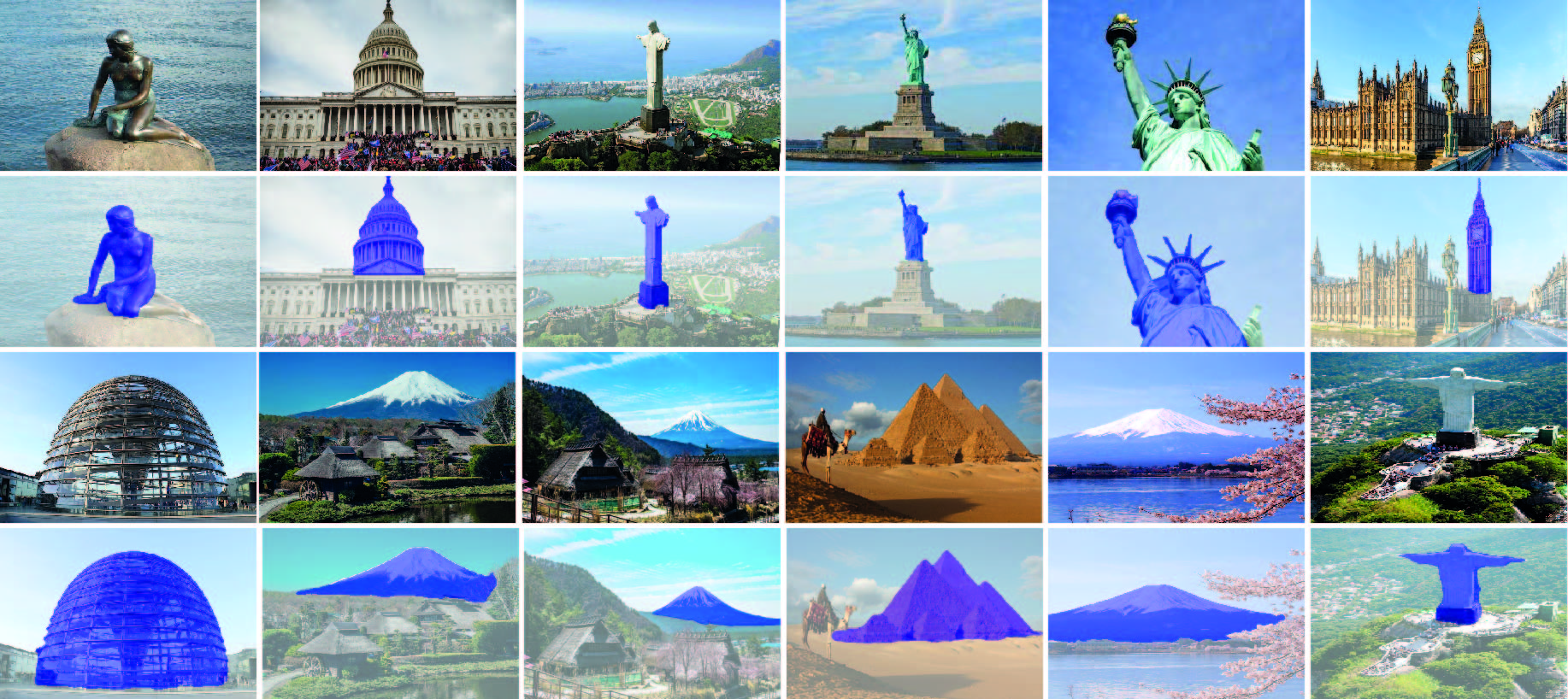}
\end{center}
\caption{
Representative examples demonstrating our segmentation performance on well-known landmarks. In the first and third rows presented the input images, while the second and fourth rows show their segmentation masks respectively.
}
\label{fig:landmark_seg}
\end{figure*}

We conducted a careful evaluation of our method against representative baselines on the ImageNet-Segmentation and PASCAL VOC 2012 datasets.  Both qualitative and quantitative experiments are presented. Additionally, we present a detailed ablation study of the design choices made by our method.

\subsection{Datasets and Evaluation Metrics}
Our evaluation is conducted on two benchmark datasets: PASCAL VOC 2012 \cite{DATASET_PACSAL_VOC_Everingham2009ThePV} validation set, and ImageNet-Segmentation\cite{DATASET_IMAGENET_SEGMENTATION_Guillaumin2014ImageNetAW}. 

\textbf{PASCAL-VOC.} A natural scene dataset, the validation set contains $1449$ images for $20$ class categories. We choose our hyperparameters using the train-set. The method is then evaluated on the validation set. 

\textbf{ImageNet-Segmentation.} The dataset consists of a subset of the ImageNet dataset. The dataset contains $4276$ images from $445$ categories.

The effectiveness of our results is demonstrated on both datasets. We also present the results of our method on images containing objects from rare class categories. Examples can be observed in \cref{fig:cool_results} and \cref{fig:landmark_seg}.


\textbf{Evaluation metric:} We evaluate our performance using mean intersection over union (mean IOU). Following \cite{Kim2020UnsupervisedLO}, mean IOU was calculated as the mean IOU of each segment in the ground truth (GT) and the estimated segment that had the largest IOU with the GT segment. Specifically, each class category was treated as an individual segment. 

\subsection{Baselines}

\textbf{Transformer Interpretability Based Segmentation (TIBS) \cite{Chefer_2021_CVPR} .} This baseline is used by our method as the atomic building block for generating relevance maps. Our method proposed multiple improvements over this baseline,  the benefit of which is evaluated. Instead of using our method, a segmentation map is generated for this baseline by thresholding its relevance map with its mean. 

\textbf{Pixel-level K-means clustering \cite{K_MEANS_MacQueen1967SomeMF}.}  
K-means clustering on RGB values of $5\times5$ patches. We use the numbers reported in \cite{Kim2020UnsupervisedLO}.\\
\textbf{Graph-based Segmentation (GS) \cite{GraphBasedSegFelzenszwalb2004EfficientGI}.}  A long-standing graph-based method. We use the reported numbers from \cite{Kim2020UnsupervisedLO}. 

\textbf{Invariant Information Clustering (IIC) \cite{Ji2019InvariantIIC} .}  
An unsupervised deep classification and segmentation method that maximizes the mutual information between different views of an image. Image segmentation is accomplished by using the IIC objective on image patches together with local spatial invariance with regard to pixel coordinates. We used the reported numbers from \cite{Kim2020UnsupervisedLO}.

\textbf{Unsupervised Segmentation (US) \cite{Kim2020UnsupervisedLO}.} This approach was described in \cref{relevanece_maps_to_segmentation}. This version of the method uses a spatial continuity loss together with a self-distillation term for feature similarity to train a network per image. It optimizes dense-per-pixel labels along with network parameters, and works within a few minutes for a  reasonable size image. The baseline does not use our relevance maps as scribble pseudo-supervision.


\begin{figure*}[t]
\begin{center}
\includegraphics[width=.99\linewidth]{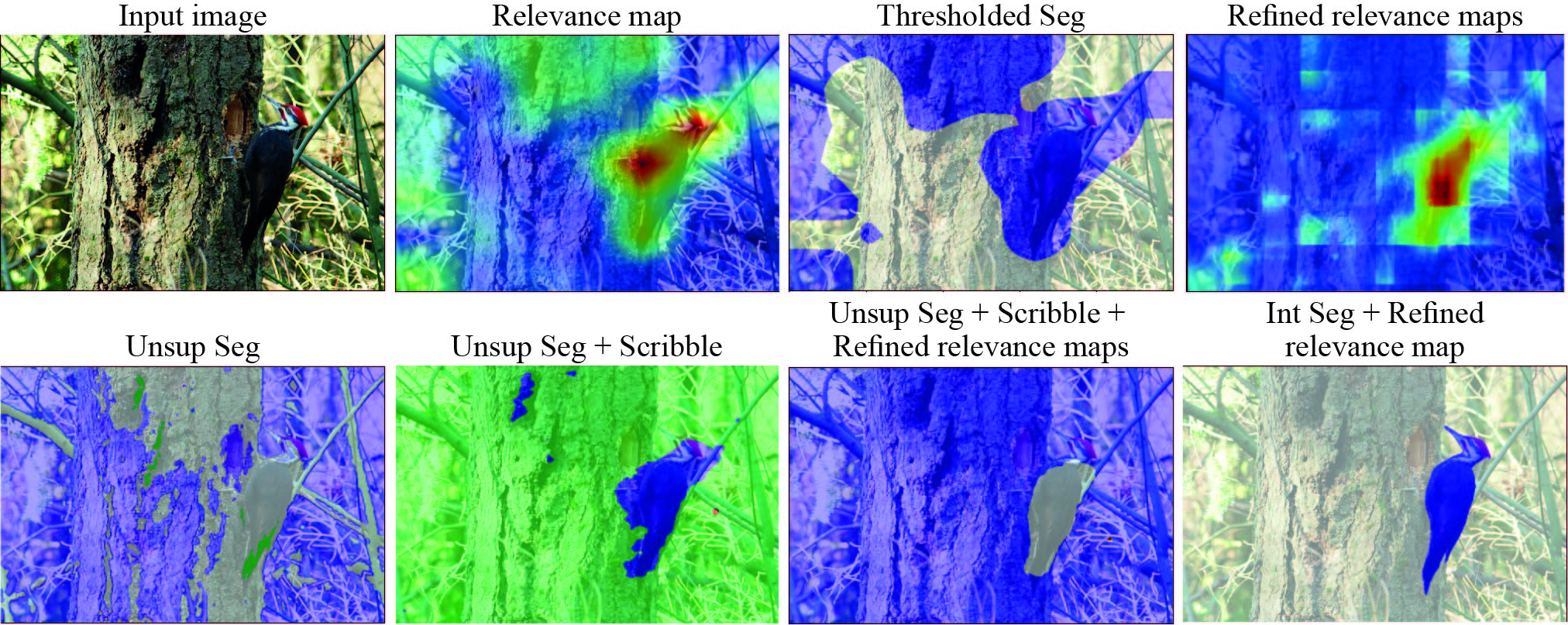}
\end{center}
\caption{
\textbf{Visual comparison on an image from PASCAL VOC.}  left to right, top row: the input image, the generated relevance map with respect to the text query "a photo of a bird", the segmentation output obtained by thresholding the relevance map, and a refined map generated using test-time augmentation of all mentioned views. bottom row: We can see that although US \cite{Kim2020UnsupervisedLO} does not perform well on its own, its combination with synthetic scribble supervision provided by our relevance maps achieved much better results. On the right, we show that using the refined relevance maps to supervise the scribble achieves even better results. Finally, we present the results of IS \cite{interactive_seg_Sofiiuk2021RevivingIT} with our refined relevance maps as pseudo-supervision, which achieves the best results.  
}
\label{fig:visual_comparison}
\end{figure*}


\subsection{Implementation Details}

We provide the main implementation details in this section. Further details can be found in the SM.

\textbf{Relevance map generation.} We use the ViT-32/B configuration of CLIP for multi-modal embedding. The relevance maps are generated using the transformer interpretability method of Chefer et al. \cite{Chefer_2021_CVPR}. For test-time augmentations we use identity, flip, contrast change, and crops. The crop parameters were crop size of $(224, 224)$, sampled on a regular grid with stride $50$. The final map was obtained by averaging all crops, using a custom probability threshold (see SM for details).   

\textbf{Segmentation generation.} We adapted two methods to use our relevance maps as pseudo-supervision: Unsupervised Clustering (UC), based on \cite{Kim2020UnsupervisedLO} and Interactive Segmentation (IS) \cite{interactive_seg_Sofiiuk2021RevivingIT}. In the configuration used, the loss weighting is as follows: continuity loss of $5.0$, feature similarity of $1.0$, and optional scribble loss of $0.5$. The configuration used for IS is HRNet-32. Further details can be found in the SM.

\begin{figure}[t]
\begin{center}
\includegraphics[width=.99\linewidth]{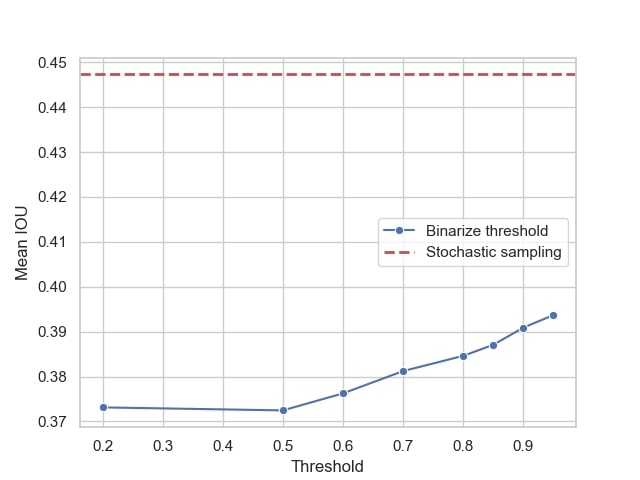}
\end{center}
\caption{\textbf{Binarization threshold analysis - }
The effect of the binarization threshold of the relevance map for a single object on segmentation of PASCAL VOC 2012 training set. The thresholded map was used without stochastic sampling as scribble supervision to the Unsupervised Clustering method.
}
\label{fig:binarization_threshold_fig}
\end{figure}
%
%
%

\section{Results}
 
 \subsection{Comparison to Other Methods}
 
 \textbf{PASCAL VOC 2012.} A quantitative evaluation is presented in \cref{tab:pascal_voc_validation_results}.  K-means performs poorly as it uses naive pixel color features. Graph-based segmentation can produce too coarse or too fine images, due to the requirement for tuning the granularity $\tau$ parameter. Like K-means, single-image IIC did not achieve strong results, and was slow.  US  did not achieve strong results on its own but achieves a $12\%$ IOU gain with our relevance maps and stochastic sampling as synthetic supervision. TIBS is essentially a thresholded interpretability map, which despite its knowledge distillation from CLIP, generated relatively noisy segmentations. Finally, we can see that the Interactive segmentation model, supervised by our method achieved the best segmentation results.

  \textbf{ImageNet-Segmentation} Results are presented in Tab.~\ref{tab:imagnet_segmentation_results}. The trends are similar to those of PASCAL VOC, our method improves over TIBS. Combination of our method with IS performs much better than with US.
  
  \begin{table}
\small
\centering
\begin{tabular}{@{}llc@{}}
\toprule
Method & Hyperparameters & Mean IOU \\
\midrule
    k-means & k=2  & 0.3166 \\
    Graph-based \cite{GraphBasedSegFelzenszwalb2004EfficientGI} & $\tau=500$  & 0.3647 \\
    IIC \cite{Ji2019InvariantIIC} & k=2  & 0.2729 \\
        
Unsup Seg \cite{Kim2020UnsupervisedLO} & Continuity loss $\mu=5$ & 0.3520  \\
    TIBS \cite{Chefer_2021_CVPR} & Mean thresholding & 0.3887 \\
    TIBS \cite{Chefer_2021_CVPR} & Clustering \cite{Kim2020UnsupervisedLO} & 0.3975 \\
    Ours + Unsup Seg \cite{Kim2020UnsupervisedLO} & Identity & 0.4867 \\
    Ours + Unsup Seg \cite{Kim2020UnsupervisedLO} & Crop  & 0.493 \\
    Ours + Unsup Seg \cite{Kim2020UnsupervisedLO} & Identity, crop   & 0.5024 \\
    Ours + Int Seg\cite{interactive_seg_Sofiiuk2021RevivingIT} & k=1, 3 clicks, crop & 0.6233 \\
    Ours + Int Seg\cite{interactive_seg_Sofiiuk2021RevivingIT} & GT k, 3 clicks, crop & \textbf{0.6392} \\
\bottomrule
\end{tabular}
\caption{
Quantitative results on PASCAL VOC 2012 validation set. k denotes the number of catgegories.
} 
\label{tab:pascal_voc_validation_results}
\end{table}
 \begin{table}
\centering
\small
\begin{tabular}{@{}llc@{}}
\toprule
Method & Hyperparameters & Mean IOU \\
\midrule
TIBS \cite{Chefer_2021_CVPR} & Mean thresholding & 0.5124 \\
TIBS \cite{Chefer_2021_CVPR} & Clustering & 0.4884 \\
    Ours + US \cite{Kim2020UnsupervisedLO} & Identity & 0.6062 \\
    Ours + IS\cite{interactive_seg_Sofiiuk2021RevivingIT} & k=1, 3 clicks, Identity & 0.6894 \\
    Ours + IS\cite{interactive_seg_Sofiiuk2021RevivingIT} & k unk., 3 clicks, Identity & 0.6908 \\
    Ours + IS\cite{interactive_seg_Sofiiuk2021RevivingIT} & k=1, 3 clicks, Crop & \textbf{0.7039} \\

\bottomrule
\end{tabular}
\caption{
Quantitative comparison on ImageNet-Segmentation. We can see that all variants of our method outperform TIBS. We can also see that combination of our method as pseudo-supervision for Interactive Segmentation is superior to combination with Unsupervised Segmentation.
} 
\label{tab:imagnet_segmentation_results}
\end{table}

 \subsection{Analysis}
 
 \textbf{Qualitative results.} We present a qualitative comparison on a single image from PASCAL VOC in \cref{fig:visual_comparison}. We observe that the refined relevance map significantly improves over the naive relevance map. We also observe that although US on its own does not generate strong results, when combined with our method, it can achieve accurate segmentations. Furthermore, we can see that combination with our refined map as pseudo-scribble supervision improves results. Finally, we see the combination of IS with our method obtains the most accurate segmentations. 
 
 \textbf{Test-time augmentations.}
 We experimented with $4$ image transformations, where each generated an independent relevance map. Overall, crops generate the most detailed relevance map. The most common approach in TTA is to aggregate predictions by averaging, to obtain a more accurate and detailed relevance map, as can be seen in \cref{tab:pascal_voc_validation_results} and \cref{fig:augmentation_reduction}, \ref{fig:visual_comparison}. We conclude that the view with the highest contribution is the crop view, but using all TTA improves segmentation performance.  
 
 \textbf{Binarization threshold and Stochastic sampling.} 
As the relevance maps already provide some segmentation signal, we provide an evaluation of direct binarization, and then passing the results as pseudo-labels to Unsupervised Clustering. We observe in \cref{fig:binarization_threshold_fig} that the threshold needs to be very high, due to the inaccuracy of the relevance map. Instead, we proposed to use stochastic sampling, which replaced the hard threshold by a softer probabilistic sampling.  We stochastically sample labels from the relevance maps in every iteration for unsupervised clustering, or by generating clicks for interactive segmentation. We found stochastic sampling to significantly improve the final quality of segmentation. 

\textbf{Unsupervised Clustering design choices.} We experimented with different relative weightings between the losses but found that the default parameters suggested by the authors showed the best results. 

 \textbf{Interactive Segmentation design choices.} The interactive segmentation method requires iterations of user clicks. Instead of using user feedback, we generated "pseudo clicks" on the fly using stochastic pixel sampling. We experimented here as well with the number of clicks generated, and found that the best results were obtained with 3 clicks. 
 
 \begin{figure}[t]
\begin{center}
\includegraphics[width=.99\linewidth]{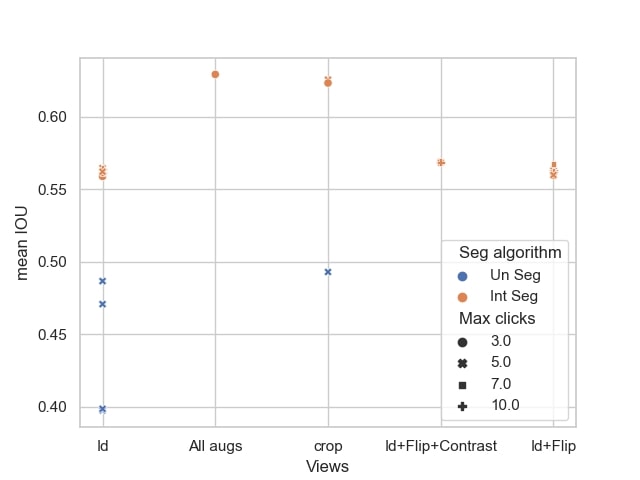}
\end{center}
\caption{
\textbf{Segmentation analysis} - In this figure, we illustrate our results from the PASCAL VOC validation set for $k=1$. Over all sets of augmentation views, interactive segmentation is superior to unsupervised segmentation. Additionally, we see that the crop view is better than using identity, flip, and contrast views together, even though using all views together is better than using the crop view alone. }
\label{fig:analysis}
\end{figure}

\section{Discussion}

\noindent \textbf{Other segmentation methods.} We presented a method for obtaining a good initial segmentation of an image in a zero-shot manner. This is then used to initialize a downstream segmentation algorithm. We showed how to combine our method with segmentation methods designed for human annotated inputs, replacing the manual annotations by our automatic method. Additionally, we showed that our method can be incorporated into an unsupervised segmentation method as extra "pseudo supervision", significantly improving performance. These two methods are just an example of the potential of the method. Our method can be incorporated with any single-image segmentation method, supervised or unsupervised. Investigating the combination with other segmentation methods e.g. GrabCut is left for future work.\\

\noindent \textbf{Augmentations.} Within the range of test-time augmentations that we investigated, we found the crop augmentations were the most significant. We believe that this augmentation can be further adapted to our task by taking crops with small step size, or of different scales in order to create relevance maps with better efficiency for objects at different scales. We did not do this in this paper due to run-time considerations.


\section{Limitations}

\noindent \textbf{Reliance on cross-modal embedding models.} Our method relies on pretrained cross-modal models trained on web data for generating the relevance maps. These models present several limitations including: being less effective on specialist domains e.g. medical imaging, not supporting other languages beyond English and Chinese. With the improvement in cross-modal embedding models, we expect these limitations to be overcome.\\

\noindent \textbf{Runtime.} Our method can be trained within a few seconds to two minutes. This is not as fast as the inference time of supervised segmentation methods, which can be done in real-time. However, our method is only a prototype and its runtime can be significantly optimized, for instance by defining improved criteria for early stopping of the clustering procedure. Additionally, computing the relevance maps for the different view can be performed in parallel which will significantly speed up results.
\section{ Conclusions}
We presented a new semantic segmentation method that can segment any object for which textual description can be provided without requiring pixel supervision or a training dataset with multiple images. We demonstrated that vision-language models can be used to transform image-level guidance into high accuracy relevance maps. These relevance maps can provide pseudo-supervision for existing single-image segmentation methods that require human annotated pixel-level supervision. We justified our design choices, and demonstrated that our qualitative and quantitative results surpass other methods requiring similar amounts of supervision. One future extension of our method is to combine the CLIP interpretability with the final segmentation stage end-to-end.

{\small
\bibliographystyle{ieee_fullname}
\bibliography{main}
}

\end{document}